# Reducing the Arity in Unbiased Black-Box Complexity

Benjamin Doerr and Carola Winzen[*]

Max-Planck-Institut für Informatik, 66123 Saarbrücken, Germany

**Abstract**

We show that for all $1 < k \leq \log n$ the $k$-ary unbiased black-box complexity of the $n$-dimensional ONEMAX function class is $O(n/k)$. This indicates that the power of higher arity operators is much stronger than what the previous $O(n/\log k)$ bound by Doerr et al. (Faster black-box algorithms through higher arity operators, Proc. of FOGA 2011, pp. 163–172, ACM, 2011) suggests.

The key to this result is an encoding strategy, which might be of independent interest. We show that, using $k$-ary unbiased variation operators only, we may simulate an unrestricted memory of size $O(2^k)$ bits.

## 1 Introduction

Black-box complexity theory tries to give a theory-driven answer to the question how difficult a problem is to be solved by general purpose optimization approaches ("black-box algorithms"). The recently introduced notion of unbiased black-box complexity in addition allows a distinction regarding the arity of the variation operators employed (see the theory track best-paper award winner by Lehre and Witt [LW10]). The only result so far indicating that there exists a hierarchy of the unbiased black-box models with respect to their arity (that is, the only result indicating that for any $k \in \mathbb{N}$ the $(k + 1)$-ary operators are strictly more powerful than $k$-ary ones) is the result of Doerr, Johannsen, Kötzing, Lehre, Wagner, and Winzen showing that the $k$-ary unbiased black-box complexity of the $n$-dimensional ONEMAX$_n$ function class is $O(n/\log k)$, $1 < k \leq n$. This function class contains for each length-$n$ bit string $z$ the function $\text{OM}_z(\cdot)$ that assigns to each bit string $x$ the number $\text{OM}_z(x) := |\{j \in [n] \mid x_j = z_j\}|$ of positions in which $x$ and $z$ agree. In this work, we given an indication that the hierarchy of complexity models might be even much more pronounced than what [DJK$^+$11] suggests. In fact, we show that for $1 < k \leq \log n$ the $k$-ary unbiased black-box complexity is $O(n/k)$. In particular, the (irrespective of the arity) best possible complexity of $\Theta(n/\log n)$ is attained already for arity $\log n$ instead of a linear arity.

We note that all other results on higher arity black-box complexities [DKLW11, DW12a, RV11] consider either only the binary or 3-ary black-box models or they consider the $*$-ary model, in which the arity of the operators may be arbitrarily large. As shown by Rowe and Vose in [RV11], for the latter model, the unbiased black-box complexity coincides with the unrestricted one by Droste, Jansen, and Wegener [DJW06].

[*]Carola Winzen is a recipient of the Google Europe Fellowship in Randomized Algorithms. This research is supported in part by this Google Fellowship.

## 1.1 Black-Box Complexity

Black-box complexity as a measure for problem difficulty was introduced by Droste, Jansen, and Wegener in their seminal paper [DJW06]. Roughly speaking, the black-box complexity is the least number of fitness evaluations needed to solve a problem. It thus is a lower bound for the performance of any evolutionary algorithm and any other randomized search heuristic.

For the ONEMAX$_n$ function class, Droste, Jansen, and Wegener proved a lower bound of $\Omega(n/\log n)$ via information theoretic arguments. That this bound is asymptotically tight was later shown by Anil and Wiegand [AW09]. As noted in [DW12b], both bounds were in fact known much earlier, in completely different fields of theoretical computer science (coin weighting games [ER63], Mastermind [Chv83]).

Droste, Jansen, and Wegener also note that their *unrestricted* version of black-box complexity sometimes gives unexpectedly low complexities. For example, the NP-complete maximum clique problem has a black-box complexity of only $O(n^2)$ as witnessed by the algorithm querying all possible two-cliques, learning all edges by this, and from this information learning the maximum clique. As possible solution they suggest adding a memory restriction. As far as we know, this line of research, however, has not created much follow-up work—with one exception being the recent result by Doerr and Winzen that even restricting the memory to a size of one does not change the black-box complexity of ONEMAX$_n$, cf. [DW12b].

A different way of restricting the class of regarded black-box algorithms was suggested by Lehre and Witt [LW10]. Noting that many randomized search heuristics treat the search space in a symmetric manner, they suggest an unbiased black-box model in which (i) new search points can only be generated from previously queried ones (or may be sampled randomly from a uniform distribution) and (ii) all variation operators used for this purpose have to be unbiased, that is, treat both bit positions and bit values in a symmetric fashion. In addition to hopefully giving more meaningful black-box complexities, this model also allows a natural definition of the arity of an algorithm: an algorithm is called $k$-ary if only variation operators are used that take up to $k$ previous search points as input. So with this model, we also have a complexity theoretic tool to discuss the role of, say, crossover in evolutionary computation.

In [DKLW11] and [RV11] different generalizations of the unbiased black-box model can be found. The model by Lehre and Witt works only for pseudo-Boolean functions, whereas Rowe and Vose [RV11] present a framework for general function classes and Doerr et al. [DKLW11] consider different unbiasedness notions for the single-source shortest paths problem.

While in [LW10] only 1-ary ("unary") variation operators were regarded, the work [DJK$^+$11] gives the first result indicating an increasing strength of higher arity operators. For the $n$-dimensional ONEMAX$_n$ function class, they show an upper bound for the $k$-ary unbiased black-box complexity of $O(n/\log k)$, $2 \leq k \leq n$. This bound is sharp for $k = n^{\Omega(n)}$. Together with the $\Theta(n \log n)$ bound for the unary case [LW10], this proves an advantage of using higher arity operators. While no lower bounds (apart from the cases $k = 1$ and $k = n^{\Omega(n)}$) are known, the upper bound of $O(n/\log k)$ does not indicate huge gains from using variation operators of arity higher than 2.

## 1.2 Our Result

In this work, we show that the increasing power of higher arities may be much more pronounced than what seems to have been known so far. We prove that, for all $2 \leq k \leq \log n$, the $k$-ary unbiased black-box complexity is at most $O(n/k)$, replacing the previous logarithmic gain by a linear one. The proof of this result uses a number of arguments that might be of interest beyond the mere result. In particular, we design $(\kappa+2)$-ary operators that allow to simulate a memory of $2^\kappa$ bits in an unrestricted fashion (*encoding technique*). Also, to save memory space, we apply a derandomized version of the random sampling strategies used in the three papers proving the $O(n/\log n)$ upper bound for the unrestricted black-box complexity of $\text{OneMax}_n$. This derandomized result says that for any $n \in \mathbb{N}$ there exists a fixed sequence of $O(n/\log n)$ search points such that querying their fitness with respect to *any* $\text{OneMax}_n$-function *surely* reveals the optimum of this function. The result itself was proven in [ER63]. However, our result seems to be the first application of the derandomized version in black-box optimization.

We do give in that we do not have any new lower bounds for the $k$-ary black-box complexity of $\text{OneMax}_n$. In particular, we cannot rule out that already for $k=2$ the unbiased black-box complexity is $\Theta(n/\log n)$. Unfortunately, currently no such bounds are known. This remains a significant open problem for which the current lower bound methods all seem to fail.

## 2 The Unbiased Black-Box Model

As mentioned in the introduction, some shortcomings of the unrestricted black-box model inspired Lehre and Witt [LW10] to introduce a new, more restrictive black-box model. This is the *unbiased black-box model*. It is based on the observation that many randomized search heuristics employ only such variation operators that are *unbiased*. In this section, we briefly introduce the unbiased black-box model and we define unbiased black-box complexity.

Roughly speaking, an unbiased variation operator must treat the bit positions $1,\ldots,n$ and the bit entries 0 and 1 in an unbiased ("fair") way. In particular, an unbiased operator may not require a specific bit value to be set to 0 or to 1. As mentioned, the unbiased model, in addition to excluding some highly artificial algorithms, also admits a notion of arity. A $k$-ary unbiased black-box algorithm is one that employs only such variation operators that take up to $k$ arguments.

For a formal definition of the unbiased black-box model let us briefly fix some notation. For all positive integers $k \in \mathbb{N}$ we abbreviate $[k] := \{1,\ldots,k\}$ and $[0..k] := [k] \cup \{0\}$. By $e_k^n$ we denote the $k$-th unit vector $(0,\ldots,0,1,0,\ldots,0)$ of length $n$. The bitwise exclusive-or is denoted by $\oplus$. For $n \in \mathbb{N}$, by $S_n$ we denote the set of all permutations of $[n]$. For $\pi \in S_n$ and $x \in \{0,1\}^n$ we abbreviate $\pi(x) := x_{\pi(1)} \ldots x_{\pi(n)}$. All logarithms log are binary logarithms.

**Definition 1** ($k$-ary unbiased operator). *Let $k \in \mathbb{N}$. A $k$-ary unbiased distribution $(\mathcal{D}(.\mid y^{(1)},\ldots,y^{(k)}))_{y^{(1)},\ldots,y^{(k)} \in \{0,1\}^n}$ is a family of probability distributions over $\{0,1\}^n$ such that for all inputs $y^{(1)},\ldots,y^{(k)} \in \{0,1\}^n$ the following two conditions hold.*

$$(i)\, \forall x, z \in \{0,1\}^n : \mathcal{D}(x \mid y^{(1)},\ldots,y^{(k)}) = \mathcal{D}(x \oplus z \mid y^{(1)} \oplus z,\ldots,y^{(k)} \oplus z),$$
$$(ii)\, \forall x \in \{0,1\}^n \, \forall \pi \in S_n : \mathcal{D}(x \mid y^{(1)},\ldots,y^{(k)}) = \mathcal{D}(\pi(x) \mid \pi(y^{(1)}),\ldots,\pi(y^{(k)})).$$

*We refer to the first condition as $\oplus$-invariance and we refer to the second as permutation invariance. A variation operator creating an offspring by sampling from a $k$-ary unbiased distribution is called a $k$-ary unbiased variation operator.*

Note that the combination of $\oplus$- and permutation invariance can be characterized as invariance under Hamming-automorphisms: $\mathcal{D}(\cdot \mid \cdot,\ldots,\cdot)$ is unbiased if and only if, for all $\alpha : \{0,1\}^n \to \{0,1\}^n$

**Algorithm 1:** Scheme of a $k$-ary unbiased black-box algorithm

1. **Initialization:** Sample $x^{(0)} \in \{0,1\}^n$ uniformly at random and query $f(x^{(0)})$;
2. **Optimization: for** $t = 1, 2, 3, \ldots$ **do**
3.     Depending on $(f(x^{(0)}), \ldots, f(x^{(t-1)}))$ choose $k$ indices $i_1, \ldots, i_k \in [0..t-1]$ and a $k$-ary unbiased distribution $(\mathcal{D}(. \mid y^{(1)}, \ldots, y^{(k)}))_{y^{(1)}, \ldots, y^{(k)} \in \{0,1\}^n}$;
4.     Sample $x^{(t)}$ according to $\mathcal{D}(. \mid x^{(i_1)}, \ldots, x^{(i_k)})$ and query $f(x^{(t)})$;

preserving the Hamming distance (i.e., $|\{j \in [n] \mid x_j = y_j\}| = |\{j \in [n] \mid (\alpha(x))_j = (\alpha(y))_j\}|$ for all $x, y$) and all bit strings $y^{(1)}, \ldots, y^{(k)}, x \in \{0,1\}^n$ we have $\mathcal{D}(x \mid y^{(1)}, \ldots, y^{(k)}) = \mathcal{D}(\alpha(x) \mid \alpha(y^{(1)}), \ldots, \alpha(y^{(k)}))$).

1-ary operators, also called *unary* operators, are often referred to as *mutation operators*. In fact, the standard bitwise mutation operator (as used, e.g., by many evolutionary algorithms) is a unary unbiased variation operator.

2-ary operators, also called *binary* operators, are typically referred to as *crossover operators*. The classic *uniform crossover operator* is an unbiased binary one. Given two search points $x$ and $y$, the uniform crossover operator creates an offspring $z$ from $x$ and $y$ by choosing independently for each index $i \in [n]$ the entry $z_i \in \{x_i, y_i\}$ uniformly at random. However, the standard *one-point crossover operator*—which, given two search points $x, y \in \{0,1\}^n$ picks uniformly at random an index $k \in [n]$ and creates from $x$ and $y$ the two offsprings $x' := x_1 \ldots x_k y_{k+1} \ldots y_n$ and $y' := y_1 \ldots y_k x_{k+1} \ldots x_n$—is not permutation-invariant, and hence not an unbiased operator.

A $k$-ary unbiased black-box algorithm is an algorithm following the scheme of Algorithm 1.

## 2.1 Unbiased Black-Box Complexity

Note that Algorithm 1 runs forever. This is justified by the fact that as performance measure of a black-box algorithm we take the number of queries performed by the algorithm until it first queries an optimal solution ("first hitting time"). More precisely, we consider the expected number of such queries as we regard random algorithms. This is the standard performance measure for randomized search heuristics, because in typical applications of such heuristics, evaluating the fitness of the search points is more costly than the generation of new ones.

Formally, for a $k$-ary unbiased black-box algorithm $A$ and a function $f : \{0,1\}^n \to \mathbb{R}$, let $T(A, f) \in \mathbb{R} \cup \{\infty\}$ be the expected number of fitness evaluations until $A$ queries for the first time some $x \in \arg\max f$. We call $T(A, f)$ the *runtime of $A$ for $f$*.

Following the usual approach in complexity theory, for a class $\mathcal{F}$ of functions $\{0,1\}^n \to \mathbb{R}$, the *$A$-black-box complexity* of $\mathcal{F}$ is $T(A, \mathcal{F}) := \sup_{f \in \mathcal{F}} T(A, f)$, the worst-case runtime of $A$ on $\mathcal{F}$.

The *$k$-ary unbiased black-box complexity* of $\mathcal{F}$ is $T(\mathcal{A}, \mathcal{F}) := \inf_{A \in \mathcal{A}} T(A, \mathcal{F})$, the minimum ("best") complexity among all $A \in \mathcal{A}$ for $\mathcal{F}$, where we denote by $\mathcal{A}$ the class of all $k$-ary unbiased black-box algorithms for the functions in $\mathcal{F}$. That is, the $k$-ary unbiased black-box complexity of some class of functions $\mathcal{F}$ is the complexity of $\mathcal{F}$ with respect to all $k$-ary unbiased black-box algorithms.

## 3 The Black-Box Complexity of OneMax

One test function often regarded in the randomized search heuristics community is the so-called "OneMax" function OM, which simply counts the number of 1-bits, $\text{OM}(x) = \sum_{i=1}^{n} x_i$. It is a linear pseudo-Boolean function with all bit weights set to one. Therefore, we consider OM one of the simplest linear pseudo-Boolean functions.

Much theoretical work has been done for this function. In fact, already one of the first theoretical works on the $(1+1)$ evolutionary algorithm (EA) [Müh92] studies the runtime of this algorithm on OM. In [DJW10] it has been shown that, in fact, OM is not only considered one of the simplest linear pseudo-Boolean functions but that the runtime of the (1+1) EA on OM is at most as large as it is on any pseudo-Boolean function with a unique global optimum.

The natural generalization of OM to a non-trivial class of functions is as follows.

**Definition 2** (ONEMAX$_n$). *Let $n \in \mathbb{N}$. For $z \in \{0,1\}^n$ let*

$$\text{OM}_z : \{0,1\}^n \to [0..n], x \mapsto |\{j \in [n] \mid x_j = z_j\}|,$$

*the function that counts the number of positions in which $x$ and $z$ agree. The string $z = \arg\max \text{OM}_z$ is called the* target string *of $\text{OM}_z$.*

*Let $\text{ONEMAX}_n := \{\text{OM}_z \mid z \in \{0,1\}^n\}$ be the set of all generalized* ONEMAX *functions.*

The ONEMAX$_n$ function class has been studied in several contexts. For example, Erdős and Rényi studied it already in the sixties in the context of coin-weighing problems [ER63]. They show that $t = (1 + o(1))(2n/\log n)$ queries to $\text{OM}_z$ are necessary and sufficient to determine a fixed (unknown!) target string $z$, with high probability.

The results of Erdős and Rényi were rediscovered by Chvátal [Chv83] in his studies on the Mastermind problem. Chvátal generalizes the upper bound given in [ER63] and shows that for any constant number $k \in \mathbb{N}$, $\Theta(n/\log n)$ queries are sufficient to identify any $k$-color string $z \in [k]^n$. In his Mastermind model, queries are $k$-color strings $x \in [k]^n$ and the functions that needs to be optimized are $\text{OM}_z : [k]^n \to [0..n], x \mapsto |\{j \in [n] \mid x_j = z_j\}|$, $z \in [k]^n$.

Doerr and Winzen [DW12b] show that the memory-1 restricted black-box complexity of ONEMAX$_n$ is $\Theta(n/\log n)$, disproving an earlier conjecture by Droste, Jansen, and Wegener [DJW06].

As mentioned, the results in [ER63] and [Chv83] seem to have been overlooked in the randomized search heuristics community for several years. For this reason, the lower bound of $\Omega(n/\log n)$ was rediscovered by Droste, Jansen, and Wegener in [DJW06, Theorem 4] and the upper bound of $(1 + o(1))2n/\log n$ was also proven by Anil and Wiegand in [AW09].

For the unary unbiased black-box model Lehre and Witt [LW10] show that the complexity of ONEMAX$_n$ is $\Theta(n \log n)$. For arities $2 \leq k \leq n$ the best known bounds are $O(n/\log k)$ proven in [DJK$^+$11]. We improve these bounds in the following theorem, which is the main result of this paper.

**Theorem 3.** *Let $n \in \mathbb{N}$ and let $1 < k \leq \log n$. The $k$-ary unbiased black-box complexity of* ONEMAX$_n$ *is $O(n/k)$.*

For constant values $k$, the result from [DJK$^+$11] shows that the $k$-ary unbiased black-box complexity of ONEMAX$_n$ is at most linear in $n$. Therefore, we need to show Theorem 3 for non-constant $k$, i.e., for $k = \omega(1)$.

## 3.1 Random Sampling

The basis for most results on the black-box complexity of $\text{OneMax}_n$ is the *random sampling technique*, which was discovered in the already mentioned work by Erdős and Rényi [ER63]. They show that if we take $t = \Theta(n/\log n)$ queries $x^1, \ldots, x^t$ from $\{0,1\}^n$ independently and uniformly at random, then, with high probability, only one possible target string remains. That is, they show that the set of all feasible target strings

$$\mathcal{S}_{\text{feas}}(z) := \mathcal{S}_{\text{feas}}(z, x^1, \ldots, x^t) := \{y \in \{0,1\}^n \mid \forall i \in [t] : \text{Om}_z(x^i) = \text{Om}_y(x^i)\}$$

has size 1, with high probability.

---
**Algorithm 2:** The random sampling technique for maximizing $\text{OneMax}_n$.

**1 Initialization:** $t \leftarrow \left\lceil \left(1 + \frac{4 \log \log n}{\log n}\right) \frac{2n}{\log n} \right\rceil$;
**2 repeat**
**3**     **for** $i = 1, \ldots, t$ **do**
**4**        Choose $x^i$ from $\{0,1\}^n$ uniformly at random and query $\text{Om}_z(x^i)$;
**5**     Set $y \leftarrow \text{chooseConsistent}(x^1, \ldots, x^t)$;
**6**     Query $\text{Om}_z(y)$;
**7 until** $\text{Om}_z(y) = n$;

---

In [DJK+11] it has been shown that this random sampling technique, for which we give its pseudo-code in Algorithm 2, can also be done unbiasedly. To this end, one has to show that the operator $\text{chooseConsistent}(\cdot, \ldots, \cdot)$, which given $t$ bit strings $x^1, \ldots, x^t \in \{0,1\}^n$ samples from $\mathcal{S}_{\text{feas}}(z, x^1, \ldots, x^t)$ uniformly at random, is an unbiased one.

The random sampling technique can be formalized as follows.

**Theorem 4** (Random Sampling Technique). *Let $n \in \mathbb{N}$ and let $z \in \{0,1\}^n$. Set $t := \lceil (1 + \frac{4 \log \log n}{\log n}) \frac{2n}{\log n} \rceil = (1 + o(1))2n/\log n$. If $x^1, \ldots, x^t$ are sampled from $\{0,1\}^n$ independently and uniformly at random, then with probability at least $1 - o(1)$ the set $\mathcal{S}_{feas}(z, x^1, \ldots, x^t)$ contains only the target string $z$ itself. That is, for all $y \in \{0,1\}^n \setminus \{z\}$ there exists an index $i \in [t]$ such that $\text{Om}_z(x^i) \neq \text{Om}_y(x^i)$.*

## 3.2 Derandomized Random Sampling

For the improved upper bound on the $k$-ary unbiased black-box complexity of $\text{OneMax}_n$, we will need a derandomized version of the random sampling technique. This is what we describe next.

**Definition 5.** *Let $n, t \in \mathbb{N}$. A sequence $x^1, \ldots, x^t \in \{0,1\}^n$ is string-distinguishing if for any two length-$n$ bit strings $y, z \in \{0,1\}^n$ with $y \neq z$ there exists an index $i = i(y, z) \in [t]$ such that $\text{Om}_z(x^i) \neq \text{Om}_y(x^i)$.*

Put differently, the sequence $x^1, \ldots, x^t \in \{0,1\}^n$ is string-distinguishing if for all strings $z \in \{0,1\}^n$ the set $\mathcal{S}_{\text{feas}}(z)$ contains only $z$.

The problem of finding string-distinguishing sequences was also considered in the work by Erdős and Rényi [ER63]. In fact, they prove the following statement.

**Theorem 6** (Derandomized Random Sampling [ER63]). *For any $\delta > 0$ there exists a positive integer $n_0 \in \mathbb{N}$ such that for all $n \geq n_0$ there exists a string-distinguishing sequence $r^1, \ldots, r^t \in \{0,1\}^n$ with $t = (1+\delta)\log(9)n/\log(n)$.*

The proof of Theorem 6 uses the probabilistic method and hence, is a non-constructive one. That is, although we know that such string-distinguishing sequences of length $t$ exist, the proof by Erdős and Rényi does not show how to explicitely construct such a sequence. However, this does not matter to us: Since in black-box complexity we count only the number of queries needed by the "best" algorithm to optimize $\textsc{OneMax}_n$, we may assume that our algorithms, given the problem size $n$, first do some preprocessing to compute a string-distinguishing sequence $r^1, \ldots, r^t$. This can be done deterministically. The preprocessing itself does not require any query, because the algorithm can check without querying anything, whether for any two bit strings $y, z \in \{0,1\}^n$ with $y \neq z$ there is at least one index $i \in [t]$ such that $\textsc{Om}_z(r^i) \neq \textsc{Om}_y(r^i)$.

To conclude this section, we briefly note that $\log(9)$ is roughly 3.17. Since we are not interested in constant factors here in this work, we settle for the fact that, for large enough $n$, string-distinguishing sequences of length $t = 3.5n/\log n$ exist.

### 3.3 Invariance of String-Distinguishing Sequences

Since we are dealing in this paper with unbiased black-box complexities, we need to ensure that we employ only unbiased variation operators. In the unbiased model we can (typically) not access particular bit positions nor can we require a particular bit value to be set to 1 or to be set 0, respectively. However, we can work around this technical difficulty. One ingredient of this work-around is the observation that the set of string-distinguishing sequences is invariant under Hamming-automorphisms.

**Theorem 7** (Invariance Property). *Let $n \in \mathbb{N}$ be large, let $t = 3.5n/\log n$, and let $r^1, \ldots, r^t$ be a string-distinguishing sequence. For all $w \in \{0,1\}^n$ and all permutations $\pi$ of $[n]$ the sequence $\pi(r^1 \oplus w), \ldots, \pi(r^t \oplus w)$ is string-distinguishing.*

*Proof.* Let $w \in \{0,1\}^n$ let $\pi$ be a permutations of $[n]$. Let $y, z \in \{0,1\}^n$. Since $r^1, \ldots, r^t$ is string-distinguishing, there exists an index $i \in [t]$ such that
$$\textsc{Om}_{\pi^{-1}(z) \oplus w}(r^i) \neq \textsc{Om}_{\pi^{-1}(y) \oplus w}(r^i).$$

We show that
$$\textsc{Om}_z(\pi(r^i \oplus w)) \neq \textsc{Om}_y(\pi(r^i \oplus w)).$$

Observe that for all $j \in [n]$ we have $(\pi^{-1}(z) \oplus w)_j = r^i_j$ if and only if $z_{\pi^{-1}(j)} = (\pi^{-1}(z))_j = (r^i \oplus w)_j = (\pi(r^i \oplus w))_{\pi^{-1}(j)}$. From this we conclude
$$\textsc{Om}_z(\pi(r^i \oplus w)) = \textsc{Om}_{\pi^{-1}(z) \oplus w}(r^i)$$
$$\neq \textsc{Om}_{\pi^{-1}(y) \oplus w}(r^i) = \textsc{Om}_y(\pi(r^i \oplus w)).$$

□

# 4 The Encoding Technique

The main contribution of this work is an *encoding technique* which we present in this section. We show that using only $k$-ary variation operators, we can create a storage of size $\Omega(2^k)$. This storage can be used to encode information. Here in our application of maximizing $\text{OneMax}_n$ functions we shall use this storage to encode fitness values $\text{Om}_z(x)$. As was done in [DJK+11], we identify $z$ by repeatedly identifying parts of $z$. This is the block-wise optimization technique which we briefly summarize below. The details of the encoding technique are presented in Sections 4.2 and 4.3.

## 4.1 Block-Wise Optimization

The paper [DJK+11] introduced a block-wise optimization technique for $\text{OneMax}_n$. We say that we "optimize" a block of length $\ell$ if we identify the entries $z_{i_1}, \ldots, z_{i_\ell}$ for some subset $\{i_1, \ldots, i_\ell\} \subseteq [n]$ of size $\ell$. The main idea of the block-wise optimization technique is to split the full length-$n$ bit string into blocks of length $\ell$ and to sequentially optimize these blocks. Each of these length-$\ell$ blocks is optimized in $O(\ell/\log \ell)$ queries via the random sampling technique that we have presented in the previous section. Since there are $\lceil n/\ell \rceil$ such length-$\ell$ blocks, the total number of queries needed by this algorithm is $O(n/\log \ell)$.

Here in this paper we shall also apply block-wise optimization. For fixed arity $k$, we set $\kappa := k - 7$ and $\ell := 2^\kappa$. We show how to optimize blocks of length $\ell$. Before we present the details, let us note already here that while $\ell$ is of order $\Theta(k)$ in [DJK+11] (thus yielding an $O(n/\log k)$ upper bound for the $k$-ary unbiased black-box complexity of $\text{OneMax}_n$), our choice of $\ell = \Omega(2^k)$ yields the claimed $O(n/k)$ upper bound.

Assume for the moment that we have created two bit strings $x, y \in \{0, 1\}^n$ such that for all bit positions in $A(x, y) := \{j \in [n] \mid x_j = y_j\}$, in which $x$ and $y$ agree, we know that $x_j = z_j$. We call the positions in $A(x, y)$ "optimized". The idea is that we need to care only about the positions $D(x, y) := [n] \setminus A(x, y)$, in which the two strings disagree. The strings $x$ and $y$ are a very condense way of encoding the information which bit positions are optimized and which are not. This encoding has been introduced in [DJK+11]. We show how to create such strings later, but—as will be discussed below—in the beginning we may simply set $y := \bar{x}$.

In the following we first describe how we create from $x$ and $y$ a storage of size $4\ell = 2^{\kappa+2}$. We then show how this storage can be used for optimizing a block of length $\ell$. As mentioned, this is the core idea in the proof of Theorem 3 and we refer to it as the *encoding technique*.

## 4.2 Creating the Storage

We describe an unbiased algorithm that, given $x$ and $y$, creates $\kappa + 3$ bit strings $y^0, \ldots, y^{\kappa+2}$ such that we can access $4\ell$ bits of $x$ by an unbiased operator.

First we need to identify the $4\ell$ bits which we want to use for storing. We define a binary operator $\texttt{findStorage}(\cdot, \cdot)$, which, given $x$ and $y$, creates from $x$ a string $y^0$ such that $D(x, y^0)$ has size $4\ell$. That is, our storage space of size $4\ell$ will be encoded in $D(x, y^0)$. We require that at least $\min\{\ell, |D(x, y)|\}$ bits of the storage are taken from the set $D(x, y)$ of not yet optimized bits. To this end, we first choose uniformly at random from $D(x, y)$ a subset $B'$ of size $\min\{\ell, |D(x, y)|\}$. Think of $B'$ as the length-$\ell$ block we aim at optimizing. Next we choose uniformly at random from $[n] \setminus B'$ a subset $S$ of size $4\ell - |B'|$. For the moment, we are not interested in optimizing the bits in

$S$ but rather need these bits for storing information. We set

$$y^0 \leftarrow \texttt{findStorage}(x,y) := x \oplus \sum_{j \in B' \cup S} e_j^n,$$

that is we create $y^0$ from $x$ and $y$ by flipping in $x$ the positions in $B'$ and $S$. The probability distribution defined by $\texttt{findStorage}(\cdot,\cdot)$ is an unbiased distribution. We omit a detailed verification of the unbiasedness of this distribution. The interested reader can find the unbiasedness proofs for all operators used in this work in the appendix.

By construction we have $D(x,y^0) = B' \cup S$, and $B' \subseteq D(x,y^0) \cap D(x,y)$. Therefore, the size of $D(x,y^0) \cap D(x,y)$ is at least $\min\{\ell, |D(x,y)|\}$. We shall need this observation later.

In this section we only allow to flip bits in $D := D(x,y^0)$. Therefore, we set

$$\mathcal{P} := \mathcal{P}(x,y^0) := \{w \in \{0,1\}^n \mid \forall j \notin D : w_j = x_j\},$$

the set of *possibly to be queried* bit strings.

For all $w \in \mathcal{P}$ we define

$$F^{(1)}(x,y^0,w) := \{j \in D \mid w_j \neq x_j\} \text{ and}$$
$$F^{(0)}(x,y^0,w) := D \backslash F^{(1)}(x,y^0,w).$$

Furthermore, we set

$$Z(x,y,y^0) := \{w \in \mathcal{P} \mid |F^{(1)}(x,y^0,w)| = 2\ell, |F^{(0)}(x,y^0,w) \cap D(x,y)| \geq \min\{\ell, |D(x,y)|\}\}.$$

The set $Z(x,y,y^0)$ contains all the strings in $\mathcal{P}$ which can be created from $x$ by flipping exactly $2\ell$ bits, keeping at least $\min\{\ell, |D(x,y)|\}$ bits in $D(x,y)$ constant.

We sample $y^1 \in Z(x,y,y^0)$ uniformly at random. This can be realized via a 3-ary unbiased operator (details are given in the appendix).

Assume that we have sampled $y^0, \ldots, y^s \in \mathcal{P}$ for some $s \in \mathbb{N}$. We describe how to sample $y^{s+1}$. For all $w \in \mathcal{P}$ and for all vectors $i \in \{0,1\}^s$ we set

$$F^{(i,1)}(x,y^0,\ldots,y^s,w) := \{j \in F^i(x,y^0,\ldots,y^s) \mid w_j \neq y_j^s\},$$
$$F^{(i,0)}(x,y^0,\ldots,y^s,w) := \{j \in F^i(x,y^0,\ldots,y^s) \mid w_j = y_j^s\},$$
$$Z(x,y,y^0,\ldots,y^s) := \{w \in \mathcal{P} \mid \forall i \in \{0,1\}^s : |F^{(i,1)}(x,y^0,\ldots,y^s,w)| = |F^i(x,y^0,\ldots,y^s)|/2\}.$$

That is, $w \in Z(x,y,y^0,\ldots,y^s)$ if and only if for all $i \in \{0,1\}^s$ the two strings $w$ and $y^s$ disagree in exactly half of the bits in $F^i(x,y^0,\ldots,y^s)$. Only for the case $s=1$ we need an additional requirement and we set

$$Z(x,y,y^0,y^1) := \{w \in \mathcal{P} \mid \forall i \in \{0,1\} : |F^{(i,1)}(x,y^0,y^1,w)| = \ell \text{ and}$$
$$|F^{(0,0)}(x,y^0,y^1,w) \cap D(x,y)| \geq \min\{\ell, |D(x,y)|\}\}.$$

Note that by the observation made above we have $|D \cap D(x,y)| \geq \min\{\ell, |D(x,y)|\}$ and, hence, the set $Z(x,y,y^0,y^1)$ is non-empty. The same is true for $Z(x,y,y^0,\ldots,y^s)$ with $2 \leq s \leq \kappa + 2$. We create $y^{s+1}$ by sampling uniformly at random from $Z(x,y,y^0,\ldots,y^s)$. Similarly as above this is a 4-ary unbiased variation operator for $s=2$ and, as was proven in [DW11, Section 5], it is an $(s+2)$-ary unbiased variation operator for $s \geq 3$. As mentioned above, details regarding the unbiasedness can be found in the appendix.

We stop this process once it holds for all $i \in \{0,1\}^s$ that the set $F^i(x, y^0, \ldots, y^s)$ contains exactly one element. Since the size of $D$ is $4\ell = 2^{\kappa+2}$, this is the case when $s = \kappa + 2$.

In what follows, for all $0 \leq s \leq \kappa + 2$ and all $i \in \{0,1\}^s$ we abbreviate $F^i := F^i(x, y^0, \ldots, y^s)$.

## 4.3 Accessing the Storage

We now describe how to access the storage space $D = B' \cup S$ of size $4\ell$. We first introduce a bijection between the length-$(\kappa+2)$ bit strings and the positions in $D$. To this end, we recall the definition of the so-called "binary-value" function

$$\mathrm{Bv} : \{0,1\}^{\kappa+2} \to [0..4\ell - 1], w \mapsto \sum_{j=1}^{\kappa+2} 2^{\kappa+2-j} w_j \,.$$

It is well-known that Bv is a bijection.

Since for all $i \in \{0,1\}^{\kappa+2}$ the set $F^i$ has size one and $F^i \neq F^j$ for all $i \neq j$, the function $\sigma : [4\ell] \to D$, which maps each value $i \in [4\ell]$ to the element in the set $F^{(\mathrm{Bv}^{-1}(i-1))}$, defines a bijection. Formally, $\sigma := \sigma(x, y, y^0, \ldots, y^{\kappa+2})$ depends on the random experiment described above but we omit this explicit notation for reasons of space.

Note that, by construction, we have $\sigma(i) \in D \cap D(x, y)$ for all elements $i \in [\min\{\ell, |D(x, y)|\}]$. This is exactly the reason why we had added the requirements $|F^{(0)}(x, y^0, w) \cap D(x, y)| \geq \min\{\ell, |D(x, y)|\}$ and $|F^{(0,0)}(x, y^0, y^1, w) \cap D(x, y)| \geq \min\{\ell, |D(x, y)|\}$ in the definition of $Z(x, y, y^0)$ and $Z(x, y, y^0, y^1)$, respectively. The block $B := \{\sigma(1), \ldots, \sigma(\min\{\ell, |D(x, y)|\})\}$ is the one we aim at optimizing. Note that $B = B'$ may or may not hold. Note further that $B = F^{(0,0)}(x, y^0, y^1, y^2)$.

We can now describe how to access the storage $D$.

For all length-$n$ bit strings $\tilde{x}, \tilde{y}, \tilde{y}^0, \ldots, \tilde{y}^{\kappa+2}$ let $\mathcal{E}^1(\tilde{x}, \tilde{y}, \tilde{y}^0, \ldots, \tilde{y}^{\kappa+2})$ be the event that all of the following statements are correct:

- $|D(\tilde{x}, \tilde{y}^0)| = 4\ell$,

- $|D(\tilde{x}, \tilde{y}^0) \cap D(\tilde{x}, \tilde{y})| \geq \min\{\ell, |D(\tilde{x}, \tilde{y})|\}$, and

- $\forall i \in [\kappa + 2] : \tilde{y}^i \in Z(\tilde{x}, \tilde{y}, \tilde{y}^0, \ldots, \tilde{y}^{i-1})$.

Abusing notation, for any such sequence of bit strings let $\tilde{\sigma} := \tilde{\sigma}(\tilde{x}, \tilde{y}, \tilde{y}^0, \ldots, \tilde{y}^{\kappa+2})$ be the bijection that maps each $i \in [4\ell]$ to the (unique) element in $F^{(\mathrm{Bv}^{-1}(i-1))}(\tilde{x}, \tilde{y}^0, \ldots, \tilde{y}^{\kappa+2})$.

Let further $s \in [4\ell]$, $r \in \{0,1\}^s$, $P \subseteq [4\ell]$ with $|P| = s$ and $\tilde{w} \in \{0,1\}^n$. For all $j \in [n]$ with $\tilde{\sigma}^{-1}(j) \in P$ let $\#\tilde{\sigma}^{-1}(j) \in [s]$ be the rank of element $\tilde{\sigma}^{-1}(j)$ within the increasingly ordered set $P$, i.e.,

$$\#\tilde{\sigma}^{-1}(j) := 1 + |\{p \in P \mid p < \tilde{\sigma}^{-1}(j)\}| \,.$$

Let $\mathcal{E}^2(r, P, \tilde{w}, \tilde{x}, \tilde{y}, \tilde{y}^0, \ldots, \tilde{y}^{\kappa+2})(w)$ be the event that all of the following statements hold

- event $\mathcal{E}^1(\tilde{x}, \tilde{y}, \tilde{y}^0, \ldots, \tilde{y}^{\kappa+2})$ holds,

- for all $j \in [n]$ with $\tilde{\sigma}^{-1}(j) \in P$ and $r_{\#\tilde{\sigma}^{-1}(j)} = 0$ as well as for all $j \in [n]$ with $\tilde{\sigma}^{-1}(j) \notin P$ we have $w_j = \tilde{w}_j$,

- for all $j \in [n]$ with $\tilde{\sigma}^{-1}(j) \in P$ and $r_{\#\tilde{\sigma}^{-1}(j)} = 1$ we have $w_j = 1 - \tilde{w}_j$.

We set $\mathcal{D}(\cdot \mid r, P, \tilde{w}, \tilde{x}, \tilde{y}, \tilde{y}^0, \ldots, \tilde{y}^{\kappa+2}) : \{0,1\}^n \to [0,1]$, $w \mapsto$

$$\begin{cases} 2^{-n} & \text{, if event } \mathcal{E}^1(\tilde{x}, \tilde{y}, \tilde{y}^0, \ldots, \tilde{y}^{\kappa+2}) \text{ does not hold,} \\ 1 & \text{, if event } \mathcal{E}^2(r, P, \tilde{w}, \tilde{x}, \tilde{y}, \tilde{y}^0, \ldots, \tilde{y}^{\kappa+2})(w) \text{ does hold,} \\ 0 & \text{, otherwise.} \end{cases}$$

For proving that for all $P \subseteq [4\ell]$ and all $r \in \{0,1\}^{|P|}$ the family

$$\left(\mathcal{D}(\cdot \mid r, P, \tilde{w}, \tilde{x}, \tilde{y}, \tilde{y}^0, \ldots, \tilde{y}^{\kappa+2})\right)_{\tilde{w}, \tilde{x}, \tilde{y}, \tilde{y}^0, \ldots, \tilde{y}^{\kappa+2} \in \{0,1\}^n}$$

is an unbiased distribution, observe first that, if event $\mathcal{E}^1(\tilde{x}, \tilde{y}, \tilde{y}^0, \ldots, \tilde{y}^{\kappa+2})$ holds, then the event $\mathcal{E}^2(r, P, \tilde{w}, \tilde{x}, \tilde{y}, \tilde{y}^0, \ldots, \tilde{y}^{\kappa+2})(w)$ holds for exactly one bit string $w$. This shows that the function $\mathcal{D}(\cdot \mid r, P, \tilde{w}, \tilde{x}, \tilde{y}, \tilde{y}^0, \ldots, \tilde{y}^{\kappa+2})$ defines a probability distribution on $\{0,1\}^n$.

To show the unbiasedness of this distribution, we need to show that for all permutations $\pi \in S_n$ and all strings $v, w \in \{0,1\}^n$ that

- $\mathcal{E}^1(\tilde{x}, \tilde{y}, \tilde{y}^0, \ldots, \tilde{y}^{\kappa+2})$ holds if and only if $\mathcal{E}^1(\pi(\tilde{x} \oplus v), \pi(\tilde{y} \oplus v), \pi(\tilde{y}^0 \oplus v), \ldots, \pi(\tilde{y}^{\kappa+2} \oplus v))$ holds, and that

- $\mathcal{E}^2(r, P, \tilde{w}, \tilde{x}, \tilde{y}, \tilde{y}^0, \ldots, \tilde{y}^{\kappa+2})(w)$ holds if and only if $\mathcal{E}^2(r, P, \pi(\tilde{w} \oplus v), \pi(\tilde{x} \oplus v), \pi(\tilde{y} \oplus v), \pi(\tilde{y}^0 \oplus v), \ldots, \pi(\tilde{y}^{\kappa+2} \oplus v))(\pi(w \oplus v))$ holds.

This computation is straightforward. It follows essentially from the two facts that **(i)** for all permutations $\pi \in S_n$ and all strings $\tilde{x}, \tilde{y}, v \in \{0,1\}^n$ we have

$$D(\pi(\tilde{x} \oplus v), \pi(\tilde{y} \oplus v)) = \{\pi(j) \mid j \in D(\tilde{x}, \tilde{y})\}$$

and that **(ii)**

$$\tilde{\sigma}\left(\pi(\tilde{x} \oplus v), \pi(\tilde{y} \oplus v), \pi(\tilde{y}^0 \oplus v), \ldots, \pi(\tilde{y}^{\kappa+2} \oplus v)\right) = \pi \circ \tilde{\sigma}(\tilde{x}, \tilde{y}, \tilde{y}^0, \ldots, \tilde{y}^{\kappa+2}).$$

We omit the details.

The operator $(\texttt{write}(r, P))(\tilde{x}, \tilde{y}, \tilde{y}^0, \ldots, \tilde{y}^{\kappa+2}, \tilde{w})$ is the one that samples from this $(\kappa + 6)$-ary distribution.

Let us briefly conclude that here in our context, for any set $P \subset [4\ell]$, any string $r \in \{0,1\}^{|P|}$, and any $w \in \{0,1\}^n$, the operator $(\texttt{write}(r, P))(x, y, y^0, \ldots, y^{\kappa+2}, w)$ is deterministic. That is, given $r, P, w, x, y, y^0, \ldots, y^{\kappa+2}$ it always outputs the same bit string, which we abbreviate by $\texttt{write}(r, P, w)$. The operator does so by setting, for each $j \in [n]$ the value $(\texttt{write}(r, P, w))_j$ to

$$\begin{cases} w_j & \text{, if } \sigma^{-1}(j) \in P \text{ and } r_{\#\sigma^{-1}(j)} = 0, \\ 1 - w_{\sigma(j)} & \text{, if } \sigma^{-1}(j) \in P \text{ and } r_{\#\sigma^{-1}(j)} = 1, \\ w_j & \text{, if } \sigma^{-1}(j) \notin P. \end{cases}$$

# 5 Proof of the Main Theorem

Throughout this proof, let $n \in \mathbb{N}$ be large and let $k \in [\log n] \cap \omega(1)$. Let $\ell := 2^\kappa$ for $\kappa := k - 7$. Let further $t = 3.5\ell/\log \ell$. By Theorem 6 there exists a string-distinguishing sequence $r^1, \ldots, r^t$ for bit strings of length $\ell$.

In what follows, instead of writing $\min\{\ell, |D(x,y)|\}$, for convenience, we simply write $\ell$.

## 5.1 Computing the Contribution of the $r^i$'s

As mentioned, the algorithm verifying Theorem 3 works by repeatedly optimizing blocks of length $\ell$. Let us assume that we are currently interested in optimizing block $B$, that is, we are interested in identifying the entries of $z$ in the positions $B$. We would like to do so by querying $r^1, \ldots, r^t$. However, this bears two difficulties: the length of each $r^i$ is only $\ell$ instead of $n$ and, secondly, we are only allowed to query bit strings that can be created by using unbiased variation operators. In particular, we cannot request a particular bit value to be set to 0 or to 1, respectively.

We show how to work around these obstacles. To this end, we first query a reference string encoding the current block we are interested in. This is the string

$$y^B := x \oplus \sum_{i \in B} e_i^n.$$

Creating $y^B$ can be done via the 4-ary unbiased operator which, given $x, y^0, y^1, y^2$ flips in $x$ all bits in $F^{(0,0)}(x, y^0, y^1, y^2)$ and sets $y_j^B = x_j$ for all $j \notin F^{(0,0)}(x, y^0, y^1, y^2)$.

Note that we can compute the contribution of the bits outside of $B$ from the fitness values of $x$ and $y^B$ by the equality

$$|\{j \in [n]\setminus B \mid x_j = z_j\}| = (\mathrm{OM}_z(y^B) + \mathrm{OM}_z(x) - \ell)/2\,.$$

This equation holds because the bits in $B$ contribute exactly $\ell$ to the sum $\mathrm{OM}_z(y^B) + \mathrm{OM}_z(x)$ and all other bits contribute either 0 or 2.

After querying $y^B$ we query the fitness values of $\mathtt{write}(r^1, [\ell], x), \ldots, \mathtt{write}(r^t, [\ell], x)$. By the observation made above, we can infer

$$\begin{aligned}
\Delta_B^i &:= \Delta_B(\mathtt{write}(r^i, [\ell], x)) \\
&:= |\{j \in B \mid \left(\mathtt{write}(r^i, [\ell], x)\right)_j = z_j\}| \\
&= \mathrm{OM}_z(\mathtt{write}(r^i, [\ell], x)) - (\mathrm{OM}_z(y^B) + \mathrm{OM}_z(x) - \ell)/2
\end{aligned}$$

from the fitness values $\mathrm{OM}_z(\mathtt{write}(r^i, [\ell], x)), \mathrm{OM}_z(y^B)$, and $\mathrm{OM}_z(x)$.

Clearly, $\Delta_B^i \in [0..\ell]$. Thus, $\kappa + 1$ bits are sufficient to store each such value in binary. Note that we sample $t = 3.5\ell/\log \ell = 3.5\ell/\kappa$ strings $\mathtt{write}(r^i, [\ell], x)$ in total. Thus, a total storage of size at most $3.5\ell(\kappa+1)/\kappa \le 4\ell = |D|$ is sufficient to store all values $\Delta_B^1, \ldots, \Delta_B^t$.

In what follows, for any $r \in [0..\ell]$ we write $\mathrm{Bv}_{\kappa+1}^{-1}(r)$ for the string that is created from $\mathrm{Bv}^{-1}(r)$ by removing the first entry (which equals 0). This is a length-$(\kappa+1)$ bit string.

For storing the values $\Delta_B^1, \ldots, \Delta_B^t$ we initialize $s \leftarrow x$ and then repeatedly update $s$ by applying the operator $\mathtt{write}$. More precisely, for $i$ running from 1 to $t$ we update $s$ by

$$s \leftarrow \mathtt{write}(\mathrm{Bv}_{\kappa+1}^{-1}(\Delta_B^i), \{(i-1)(\kappa+1)+1, \ldots, i(\kappa+1)\}, s)\,.$$

**Algorithm 3:** Optimizing $\text{OneMax}_n$ with unbiased variation operators.

1. **Initialization:** Sample $x \in \{0,1\}^n$ uniformly at random and query $\text{Om}_z(x)$;
2. Set $y \leftarrow \bar{x}$ and query $\text{Om}_z(y)$;
3. **for** *block* $b = 1, \ldots, \lceil n/\ell \rceil$ **do**
4.     Create a storage $D$ of size $4\ell$ (cf. Section 4);
5.     Query $\text{Om}_z(y^B)$;
6.     **for** $i = 1, \ldots, t$ **do**
7.         Query $\text{Om}_z\big(\texttt{write}(r^i, [\ell], x)\big)$; // query string-distinguishing sequence
8.     Set $s \leftarrow x$;
9.     **for** $i = 1, \ldots, t$ **do**
10.        Update $s \leftarrow \texttt{write}(\text{Bv}_{\kappa+1}^{-1}(\Delta_B^i), \{(i-1)(\kappa+1)+1, \ldots, i(\kappa+1)\}, s)$;
11.        Query $\text{Om}_z(s)$; // store fitness values
12.     **repeat**
13.        Sample $q \leftarrow \texttt{chooseConsistent}(x, y^B, s)$;
14.        Query $\text{Om}_z(q)$;
15.     **until** $\Delta_B(q) = \ell$;
16.     Update $x$ and $y$;

The resulting string $s$ encodes all values $\Delta_B^1, \ldots, \Delta_B^t$ and, as we shall see below, it allows us to regain the full guessing history. Note that the operator $\texttt{write}(\text{Bv}_{\kappa+1}^{-1}(\Delta_B^i), \{(i-1)(\kappa+1)+1, \ldots, i(\kappa+1)\})$ can be chosen by knowing only the fitness values $\text{Om}_z(x), \text{Om}_z(y^B)$, and $\text{Om}_z(\texttt{write}(r^i, [\ell], x))$. It is important to note that the strings $y^B$ and $\texttt{write}(r^i, [\ell], x)$ do not count toward the arity of the operator creating

$$\texttt{write}(\text{Bv}_{\kappa+1}^{-1}(\Delta_B^i), \{(i-1)(\kappa+1)+1, \ldots, i(\kappa+1)\}, s)$$
$$= \big(\texttt{write}(\text{Bv}_{\kappa+1}^{-1}(\Delta_B^i), \{(i-1)(\kappa+1)+1, \ldots, i(\kappa+1)\})\big)(x, y, y^0, \ldots, y^{\kappa+2}, s).$$

### 5.2 Optimizing $\text{OneMax}_n$

Algorithm 3 gives an overview of the algorithm verifying Theorem 3.

In this algorithm we make use of the operator $\texttt{chooseConsistent}(x, y^B, s)$ which we describe next. Given the strings $x, y, y^0, \ldots, y^{\kappa+2}, y^B$, and $s$ we first compute

$$\mathcal{F}_B := \big\{w \in \{0,1\}^n \mid \forall j \in A(x, y^B) : w_j = x_j \text{ and}$$
$$\forall i \in [t] : \text{Om}_z\big(\texttt{write}(r^i, [\ell], x)\big) = \text{Om}_w\big(\texttt{write}(r^i, [\ell], x)\big)\big\}.$$

The operator $\texttt{chooseConsistent}(x, y^B, s)$ samples from this set uniformly at random if it is non-empty and it outputs a random string otherwise. This is an unbiased operator. We refer to the appendix for a proof of the unbiasedness and note here only that the arity of the operator $\texttt{chooseConsistent}$ is $\kappa + 7$.

To compute $\mathcal{F}_B$, the operator $\texttt{chooseConsistent}$ needs to infer the fitness values $\text{Om}_z(\texttt{write}(r^i, [\ell], x))$. We show that this is possible by accessing only the information given in the strings $x, y, y^0, \ldots, y^{\kappa+2}, y^B, s$ and their corresponding fitness values: For any two numbers $a, b \in \mathbb{R}$ define the Kronecker symbol by setting $\delta(a, b) = 1$, if $a = b$, and $\delta(a, b) = 0$, otherwise.

The operator `chooseConsistent` can infer the value $\text{OM}_z\bigl(\text{write}(r^i, [\ell], x)\bigr)$ from the input strings by first computing

$$\Delta_B^i = \sum_{j=1}^{\kappa+1} 2^{\kappa+1-j}\bigl(1 - \delta(x_{\sigma((i-1)(\kappa+1)+j)}, s_{\sigma((i-1)(\kappa+1)+j)})\bigr)$$

and then adding to $\Delta_B^i$ the value $(\text{OM}_z(y^B) + \text{OM}_z(x) - \ell)/2$. This shows how to compute $\mathcal{F}_B$.

Note that here in our case we actually have $|\mathcal{F}_B| = 1$. This is by definition of the sequence $r^1, \ldots, r^t$ and Theorem 7. In particular we have $(\text{chooseConsistent}(x, y^B, s))_j = z_j$ for all $j \in B$, and, again by definition, we have $(\text{chooseConsistent}(x, y^B, s))_j = x_j$ for all $j \in [n]\backslash B = A(x, y^B)$.

As mentioned in the beginning of Section 4, in Line 16 of Algorithm 3 we want to update $x$ and $y$ in such a way that we preserve the invariance that $x_j = y_j$ only for such $j$ for which we know that $x_j = z_j$. To maintain this invariance, we first update $y$ by setting for each $j \in [n]$

$$y_j := \begin{cases} y_j & \text{, if } y_j^B = x_j, \\ \text{chooseConsistent}(x, y^B, s)_j & \text{, if } y_j^B \neq x_j. \end{cases}$$

Confer the appendix for a (straightforward) proof that $y$ can be updated by an unbiased variation operator. The arity of this update operator is 4 as it requires $x, y, y^B, \text{chooseConsistent}(x, y^B, s)$ as inputs.

After updating $y$, we update $x$ by just copying

$$x \leftarrow \text{chooseConsistent}(x, y^B, s).$$

### 5.3 Runtime of Algorithm 3

It remains to bound the runtime of Algorithm 3 by $O(n/k)$. To this end, we first observe that in Algorithm 3 the operator of largest arity (this is the `chooseConsistent` operator) has arity $\kappa + 7$. This shows that all variation operator are of arity at most $k$.

Next we observe that the (deterministic) number $N$ of queries that Algorithm 3 does to optimize a block of length $\ell$ (Lines 4–16) is $\kappa + 2t + 6$ (the queries are $y^0, \ldots, y^{\kappa+2}, y^B, \text{write}(r^1, [\ell], x), \ldots, \text{write}(r^t, [\ell], x)$ plus $t$ queries for updating $s$, the query `chooseConsistent` and one query for updating $y$). Now $t = 3.5\ell/\log \ell < 2^{\kappa+2}/\kappa = 2^{k-5}/(k-7)$ and, hence, $N < 2^{k-4}/(k-7) + k - 1$.

There are $\lceil n/\ell \rceil = \lceil n/2^{k-7} \rceil$ blocks of length $\ell$ and we have two additional queries in the beginning for initializing $x$ and $y$. Thus, Algorithm 3 does a total number of less than

$$\lceil n/2^{k-7} \rceil \left(\frac{2^{k-4}}{k-7} + k - 1\right) + 2$$

queries. This term is in $O(n/k)$ for $k \in [\log n] \cap \omega(1)$.

## 6 Conclusion

We have shown that the $k$-ary unbiased black-box complexity of $\text{OneMax}_n$ is at most $O(n/k)$ for $1 < k \leq \log n$, thus greatly improving the previously best known bound of $O(n/\log k)$. This results motivates further research in three directions.

An obvious question arising from our result are matching lower bounds. So far, the only lower bounds known are $\Omega(n \log n)$ for unary unbiased black-box algorithms and $\Omega(n/\log n)$ for the unrestricted case. Hence we currently cannot rule out that already the binary unbiased black-box complexity of $\textsc{OneMax}_n$ is $O(n/\log n)$. It seems that proving lower bounds stronger than those for the unrestricted setting and arities greater than one need the development of substantially new methods.

Both the previous $O(n/\log k)$ and in a stronger sense our new $O(n/k)$ upper bounds for the $k$-ary unbiased black-box complexity can be seen as an indication that using higher arity variation operators may increase the performance of randomized search heuristics. This seems to be a research problem for which still few theoretical results exist. There is a sequence of papers showing that a particular evolutionary algorithm for the all-pairs shortest path problem becomes more efficient when a natural crossover operator is used [DHK08, DT09, DJK$^+$10]. There also is a recent result [KST11] on OneMax, which, however, uses an uncommon shuffling operator that clearly is far from unbiased. The same paper, as well as classic works by Jansen and Wegener [JW02] show that crossover can be useful for jump functions, if population size, jump parameter and crossover rate are in a favorable relation. Despite these results, it seems that even for simple test functions like OneMax we do not understand crossover very well, and nothing seems to be known on non-artificial uses of higher arity operators.

Finally, the key methods to derive the superior black-box optimization algorithm presented in this work is a quite technical encoding technique that allows to simulate a storage of $2^\kappa$ bits in the unbiased black-box model. While we are optimistic that this technique will lead to some more unexpected black-box results, this elaborate technique also raises the question to what extent unbiased black-box complexities are a suitable measure for how difficult a problem is for a reasonable search heuristic. While we do feel that the different black-box models suggested in the recent past greatly enlarge our understanding of problem difficulty, it remains a challenging problem to find even better complexity notions.

# A  Unbiasedness of Variation Operators

We show that the variation operators used by Algorithm 3 are unbiased. We do that in the sequence they appear in the pseudo-code of Algorithm 3.

## A.1  Sampling $x$

In the first line of Algorithm 3 we sample $x \in \{0,1\}^n$. This is an unbiased (0-ary) operator. In fact, it is the only 0-ary unbiased variation operator on $\{0,1\}^n$.

## A.2  Sampling $y$

In the second line of Algorithm 3 we set $y$ to be the bit-wise complement of $x$. This is a unary unbiased variation operator as can be seen as follows. For all $x, w \in \{0,1\}^n$ it holds that $\overline{x \oplus w} = \bar{x} \oplus w$ and for all $x$ and for all $\pi \in S_n$ we have $\overline{\pi(x)} = \pi(\bar{x})$.

The corresponding distribution is

$$\mathcal{D}(\cdot \mid x) : \{0,1\}^n \to [0,1], w \mapsto \begin{cases} 1 & \text{, if } \forall j \in [n] : w_j = 1 - x_j, \\ 0 & \text{, otherwise.} \end{cases}$$

## A.3  Sampling $y^0$

We show that $\texttt{findStorage}(\cdot, \cdot)$ is an unbiased variation operator. To this end, we define for all bit strings $\tilde{x}, \tilde{y} \in \{0,1\}^n$ the distribution

$$\mathcal{D}(\cdot \mid \tilde{x}, \tilde{y}) : \{0,1\}^n \to [0,1], w \mapsto \begin{cases} 0 & \text{, if } w \notin \mathcal{S}(\tilde{x}, \tilde{y}), \\ |\mathcal{S}(\tilde{x}, \tilde{y})|^{-1} & \text{, if } w \in \mathcal{S}(\tilde{x}, \tilde{y}), \end{cases}$$

where we abbreviate

$$\mathcal{S}(\tilde{x}, \tilde{y}) := \{w \in \{0,1\}^n \mid |D(\tilde{x}, w)| = 4\ell\} \text{ and } |D(\tilde{x}, \tilde{y}) \cap D(\tilde{x}, w)| \geq \min\{\ell, |D(\tilde{x}, \tilde{y})|\}.$$

Clearly we have for all permutations $\pi$ of $[n]$ and all $z \in \{0,1\}^n$

$$D(\pi(\tilde{x} \oplus z), \pi(\tilde{y} \oplus z)) = \{\pi(j) \mid j \in D(\tilde{x}, \tilde{y})\} \tag{1}$$

and thus, $|D(\tilde{x}, \tilde{y})| = |D(\pi(\tilde{x} \oplus z), \pi(\tilde{y} \oplus z))|$.

Therefore, it is easily verified that for all $w \in \{0,1\}^n$ we have $w \in \mathcal{S}(\tilde{x}, \tilde{y})$ if and only if $\pi(w \oplus z) \in \mathcal{S}\big(\pi(\tilde{x} \oplus z), \pi(\tilde{y} \oplus z)\big)$. Hence,

$$\mathcal{D}(w \mid \tilde{x}, \tilde{y}) = \mathcal{D}\big(\pi(w \oplus z) \mid \pi(\tilde{x} \oplus z), \pi(\tilde{y} \oplus z)\big).$$

This shows that $(\mathcal{D}(\cdot \mid \tilde{x}, \tilde{y}))_{\tilde{x}, \tilde{y} \in \{0,1\}^n}$ is an unbiased distribution.

## A.4 Sampling $y^1, \ldots, y^{\kappa+2}$

We show that sampling $y^1$ is a 3-ary unbiased operation. The unbiasedness of sampling $y^2, \ldots, y^{\kappa+2}$ can be proved in a similar way.

Recall that we have set

$$Z(\tilde{x}, \tilde{y}, \tilde{y}^0) = \{w \in \mathcal{P} \mid |F^{(1)}(\tilde{x}, \tilde{y}^0, w)| = 2\ell, |F^{(0)}(\tilde{x}, \tilde{y}^0, w) \cap D(\tilde{x}, \tilde{y})| \geq \min\{\ell, |D(\tilde{x}, \tilde{y})|\}\}.$$

For all $\tilde{x}, \tilde{y}, \tilde{y}^0 \in \{0,1\}^n$ we define

$$\mathcal{D}(\cdot \mid \tilde{x}, \tilde{y}, \tilde{y}^0) : \{0,1\}^n \to [0,1], w \mapsto \begin{cases} 2^{-n} & \text{, if } \tilde{y}^0 \notin \mathcal{S}(\tilde{x}, \tilde{y}), \\ 0 & \text{, if } \tilde{y}^0 \in \mathcal{S}(\tilde{x}, \tilde{y}) \text{ and } w \notin Z(\tilde{x}, \tilde{y}, \tilde{y}^0), \\ |Z(\tilde{x}, \tilde{y}, \tilde{y}^0)|^{-1} & \text{, if } \tilde{y}^0 \in \mathcal{S}(\tilde{x}, \tilde{y}) \text{ and } w \in Z(\tilde{x}, \tilde{y}, \tilde{y}^0). \end{cases}$$

Using again equation (1) one easily shows that

$$w \in Z(\tilde{x}, \tilde{y}, \tilde{y}^0)$$

if and only if

$$\pi(w \oplus z) \in Z(\pi(\tilde{x} \oplus z), \pi(\tilde{y} \oplus z), \pi(\tilde{y}^0 \oplus z))$$

for all permutations $\pi \in S_n$ and all $z \in \{0,1\}^n$.

Together with the already mentioned fact that $\tilde{y}^0 \in \mathcal{S}(\tilde{x}, \tilde{y})$ if and only if $\pi(\tilde{y}^0 \oplus z) \in \mathcal{S}(\pi(\tilde{x} \oplus z), \pi(\tilde{y} \oplus z))$ this shows the unbiasedness of sampling $y^1$.

## A.5 Unbiasedness of Creating $y^B$

Creating $y^B$ is equivalent to sampling from the following distribution $\mathcal{D}(\cdot \mid \cdot, \cdot, \cdot, \cdot)$ with input $x, y^0, y^1, y^2$.

For all $\tilde{x}, \tilde{y}^0, \tilde{y}^1, \tilde{y}^2$ we let

$$\mathcal{D}(\cdot \mid \tilde{x}, \tilde{y}^0, \tilde{y}^1, \tilde{y}^2) : \{0,1\}^n \to [0,1], w \mapsto \begin{cases} 1 & \text{, if } \forall j \in F^{(0,0)}(\tilde{x}, \tilde{y}^0, \tilde{y}^1, \tilde{y}^2) : w_j = 1 - x_j \\ & \text{ and } \forall j \in [n] \backslash F^{(0,0)}(\tilde{x}, \tilde{y}^0, \tilde{y}^1, \tilde{y}^2) : w_j = x_j, \\ 0 & \text{, otherwise.} \end{cases}$$

Again it is straightforward to show that this is an unbiased distribution. It follows essentially from the fact that

$$F^{(0,0)}\big(\pi(\tilde{x} \oplus w), \pi(\tilde{y}^0 \oplus w), \pi(\tilde{y}^1 \oplus w), \pi(\tilde{y}^2 \oplus w)\big) = \{\pi(j) \mid j \in F^{(0,0)}(\tilde{x}, \tilde{y}^0, \tilde{y}^1, \tilde{y}^2)\}$$

## A.6 Unbiasedness of the Operator `write`

This operator has been described in the main text in much detail, cf. Section 4.3.

## A.7 Unbiasedness of `chooseConsistent`

Let $\tilde{x}, \tilde{y}, \tilde{y}^0, \ldots, \tilde{y}^{\kappa+2}, \tilde{y}^B, \tilde{s}$ be length-$n$ bit strings. We say that event $\mathcal{E}^3$ holds for these strings if all of the following events are true:

- $\mathcal{E}^1(\tilde{x}, \tilde{y}, \tilde{y}^0, \ldots, \tilde{y}^{\kappa+2})$,

- $\tilde{y}^B := \tilde{x} \oplus \sum_{i \in F^{(0,0)}(\tilde{x}, \tilde{y}^0, \tilde{y}^1, \tilde{y}^2)} e_i^n$,

- $\tilde{s}_j = \tilde{x}_j$ for all $j \in A(\tilde{x}, \tilde{y}^0)$.

Recall that for such strings we have defined

$$\tilde{\mathcal{F}}_B := \mathcal{F}_B(\tilde{x}, \tilde{y}, \tilde{y}^0, \ldots, \tilde{y}^{\kappa+2}, \tilde{y}^B, \tilde{s})$$
$$:= \left\{ w \in \{0,1\}^n \mid \forall j \in A(\tilde{x}, \tilde{y}^B) : w_j = \tilde{x}_j, \forall i \in [t] : f\big(\mathtt{write}(r^i, [\ell], \tilde{x})\big) = \mathrm{OM}_w\big(\mathtt{write}(r^i, [\ell], \tilde{x})\big) \right\},$$

where

$$f\big(\mathtt{write}(r^i, [\ell], \tilde{x})\big) = \sum_{j=1}^{\kappa+1} 2^{\kappa+1-j}\big(1 - \delta(\tilde{x}_{\tilde{\sigma}((i-1)(\kappa+1)+j)}, \tilde{s}_{\tilde{\sigma}((i-1)(\kappa+1)+j)})\big)$$
$$+ (\mathrm{OM}_z(\tilde{y}^B) + \mathrm{OM}_z(\tilde{x}) - \ell)/2$$

($\tilde{\sigma}$ is defined as in Section 4.3).

Let $\mathcal{D}(\cdot \mid \tilde{x}, \tilde{y}, \tilde{y}^0, \ldots, \tilde{y}^{\kappa+2}, \tilde{y}^B, \tilde{s}) : \{0,1\}^n \to [0,1], w \mapsto$

$$\begin{cases} 2^{-n} & \text{, if event } \mathcal{E}^3(\tilde{x}, \tilde{y}, \tilde{y}^0, \ldots, \tilde{y}^{\kappa+2}, \tilde{y}^B, \tilde{s}) \text{ does not hold,} \\ |\tilde{\mathcal{F}}_B|^{-1} & \text{, if event } \mathcal{E}^3(\tilde{x}, \tilde{y}, \tilde{y}^0, \ldots, \tilde{y}^{\kappa+2}, \tilde{y}^B, \tilde{s}) \text{ holds and } w \in \tilde{\mathcal{F}}_B \\ 0 & \text{, otherwise.} \end{cases}$$

The unbiasedness of this distribution follows essentially from the fact that we sample from $\tilde{\mathcal{F}}_B$ uniformly. First we observe that event $\mathcal{E}^3(\tilde{x}, \tilde{y}, \tilde{y}^0, \ldots, \tilde{y}^{\kappa+2}, \tilde{y}^B, \tilde{s})$ holds if only if for all permutations $\pi \in S_n$ and all $v \in \{0,1\}^n$ the event $\mathcal{E}^3\big(\pi(\tilde{x}) \oplus v, \pi(\tilde{y}) \oplus v, \pi(\tilde{y}^0) \oplus v, \ldots, \pi(\tilde{y}^{\kappa+2}) \oplus v, \pi(\tilde{y}^B) \oplus v, \pi(\tilde{s}) \oplus v\big)$ holds:

- For the events $\mathcal{E}^1$ we have argued for this already in Section 4.3.

- $F^{(0,0)}(\pi(\tilde{x}) \oplus v, \pi(\tilde{y}^0) \oplus v, \pi(\tilde{y}^1) \oplus v, \pi(\tilde{y}^2) \oplus v) = \pi\big(F^{(0,0)}(\tilde{x}, \tilde{y}^0, \tilde{y}^1, \tilde{y}^2)\big)$. Hence,

$$\tilde{y}^B := \tilde{x} \oplus \sum_{i \in F^{(0,0)}(\tilde{x}, \tilde{y}^0, \tilde{y}^1, \tilde{y}^2)} e_i^n$$

if and only if $\pi(\tilde{y}^B) \oplus v = \pi(\tilde{x}) \oplus v \oplus \sum_{i \in F^{(0,0)}(\pi(\tilde{x}) \oplus v, \pi(\tilde{y}^0) \oplus v, \pi(\tilde{y}^1) \oplus v, \pi(\tilde{y}^2) \oplus v)} e_i^n$,

- Clearly, $\tilde{s}_j = \tilde{x}_j$ for all $j \in A(\tilde{x}, \tilde{y}^0)$ if and only if $(\pi(\tilde{s}) \oplus v)_j = (\pi(x) \oplus v)_j$ for all $j \in A(\pi(\tilde{x}) \oplus v, \pi(\tilde{y}^0) \oplus v)$.

Therefore, we need to show that

$$w \in \tilde{\mathcal{F}}_B(\tilde{x}, \tilde{y}, \tilde{y}^0, \ldots, \tilde{y}^{\kappa+2}, \tilde{y}^B, \tilde{s}) \text{ if and only if} \qquad (2)$$
$$\pi(w) \oplus v \in \tilde{\mathcal{F}}\left(\pi(\tilde{x}) \oplus v, \pi(\tilde{y}) \oplus v, \pi(\tilde{y}^0) \oplus v, \ldots, \pi(\tilde{y}^{\kappa+2}) \oplus v, \pi(\tilde{y}^B) \oplus v, \pi(\tilde{s}) \oplus v\right).$$

This follows from the following two observations. For all $u, v, w \in \{0,1\}^n$ and all permutations $\pi \in S_n$ we have

$$\text{OM}_w(u) = \text{OM}_{\pi(w) \oplus v}(\pi(u) \oplus v).$$

Further we have

$$\tilde{\sigma}(\pi(\tilde{x}) \oplus v, \pi(\tilde{y}^0) \oplus v, \ldots, \pi(\tilde{y}^{\kappa+2}) \oplus v) = \pi \circ \tilde{\sigma}(\tilde{x}, \tilde{y}^0, \ldots, \tilde{y}^{\kappa+2}) : [4\ell] \to F^{(0,0)}(\tilde{x}, \tilde{y}^0, \tilde{y}^1, \tilde{y}^2),$$
$$i \mapsto \pi\left(\tilde{\sigma}(\tilde{x}, \tilde{y}^0, \ldots, \tilde{y}^{\kappa+2})(i)\right).$$

Using these two facts one easily verifies (2).

This shows that the distribution $\mathcal{D}(\cdot \mid \tilde{x}, \tilde{y}, \tilde{y}^0, \ldots, \tilde{y}^{\kappa+2}, \tilde{y}^B, \tilde{s})$ is unbiased. The operator $\texttt{chooseConsistent}(\tilde{x}, \tilde{y}^B, \tilde{s}) = \texttt{chooseConsistent}(\tilde{x}, \tilde{y}, \tilde{y}^0, \ldots, \tilde{y}^{\kappa+2}, \tilde{y}^B, \tilde{s})$ is the operator sampling from this distribution.

## A.8 Unbiasedness of Updating $y$

For all $\tilde{x}, \tilde{y}, \tilde{y}^B, \tilde{w}$ let

$$\mathcal{D}(\cdot \mid \tilde{x}, \tilde{y}, \tilde{y}^B, \tilde{w}) : \{0,1\}^n \to [0,1], w \mapsto \begin{cases} 1 & \text{, if } \forall j \in D(\tilde{x}, \tilde{y}^B) : w_j = \tilde{w}_j \\ & \text{and } \forall j \in [n] \setminus D(\tilde{x}, \tilde{y}^B) : w_j = \tilde{y}_j, \\ 0 & \text{, otherwise.} \end{cases}$$

By equation (1) this is easily verified to be a 4-ary unbiased distribution. For updating $y$ in Line 16 we sample from $\mathcal{D}(\cdot \mid x, y, y^B, \texttt{chooseConsistent}(x, y^B, s))$.